\documentclass[letterpaper, 10 pt, conference]{ieeeconf}
\usepackage{graphicx} % Required for the inclusion of images
\usepackage{hyperref}
\usepackage{subcaption}
\usepackage{colonequals}
\usepackage{amsmath} % Required for some math elements
\usepackage{float}
\usepackage{booktabs}
\usepackage{amsfonts}
\usepackage{amssymb}
\usepackage[dvipsnames]{xcolor}
\usepackage{siunitx}
\usepackage{ulem}
\usepackage[font=small,labelfont=bf]{caption}

\usepackage[ruled]{algorithm2e}
\makeatletter
\newcommand{\algorithmstyle}[1]{\renewcommand{\algocf@style}{#1}}
\newcommand{\removelatexerror}{\let\@latex@error\@gobble}
\newcommand{\norm}[1]{\left\lVert#1\right\rVert}
\makeatother
\usepackage[noadjust]{cite}

\IEEEoverridecommandlockouts                              % This command is only needed if 
\DeclareMathOperator{\E}{\mathbb{E}}

\title{\LARGE \bf
	Offline Learning of Counterfactual Predictions for Real-World Robotic Reinforcement Learning
}

\author{Jun Jin$^{*,\dagger}$, Daniel Graves$^{*}$, Cameron Haigh$^{*}$, Jun Luo$^{*}$ and Martin Jagersand$^{\dagger}$% <-this % stops a space
\thanks{$^{*}$Noah's Ark Lab, Huawei Technologies Canada, Ltd., Edmonton AB., Canada,	{\tt\small \{jun.jin1, daniel.graves, cameron.haigh, jun.luo1\}@huawei.com}.}%
	\thanks{$^{\dagger}$Department of Computing Science,
		University of Alberta, Edmonton AB., Canada
		{\tt\small \{jjin5, mj7\}@ualberta.ca}}%
}

\begin{document}
	\maketitle
	\thispagestyle{empty}
	\pagestyle{empty}
	%%%%%%%%%%%%%%%%%%%%%%%%%%%%%%%%%%%%%%%%%%%%%%%%%%%%%%%%%%%%%%%%%%%%%%%%%%%%%%%%
	\begin{abstract}
		We consider real-world reinforcement learning (RL) of robotic manipulation tasks that involve both visuomotor skills and contact-rich skills. We aim to train a policy that maps multimodal sensory observations (vision and force) to a manipulator’s joint velocities under practical considerations. We propose to use offline samples to learn a set of general value functions (GVFs) that make counterfactual predictions from the visual inputs. We show that combining the offline learned counterfactual predictions with force feedbacks in online policy learning allows efficient reinforcement learning given only a terminal (success/failure) reward. We argue that the learned counterfactual predictions form a compact and informative representation that enables sample efficiency and provides auxiliary reward signals that guide online explorations towards contact-rich states. Various experiments in simulation and real-world settings were performed for evaluation. Recordings of the real-world robot training can be found via https://sites.google.com/view/realrl.
	\end{abstract}
	
	%%%%%%%%%%%%%%%%%%%%%%%%%%%%%%%%%%%%%%%%%%%%%%%%%%%%%%%%%%%%%%%%%%%%%%%%%%%%%%%%
	\section{Introduction}
	\label{sec:intro}

We consider real-world reinforcement learning (RL) of robotic manipulation tasks that involve both visuomotor skills and contact-rich skills. We aim to train a policy that maps multimodal sensory observations (vision and force) to a manipulator’s joint velocities under the following practical considerations. (1) The training needs to be done in a reasonable amount of time. (2) Only a terminal reward (success/failure) that defines the task is given. The first consideration relates to the sample efficiency problem~\cite{zhu2020ingredients}. And the second will remove the need for reward shaping and tediously designed perception pipelines (such as object pose estimation) or calibrated instrumentation (such as optical trackers) to provide dense reward signals. Thus our problem setting matches well with real-world robotic applications. Meanwhile, our problem setting is also challenging since reinforcement learning with high-dimensional states (images) and continuous actions (joint velocities) typically requires massive amount of robot-environment interactions. Furthermore,learning with only a terminal reward remains a challenging problem.  

We address the challenges from a predictive learning perspective~\cite{pednault2000representation} which encodes high-dimensional image states as action-oriented predictions that form a compact state representation and guide online explorations towards contact-rich states. In cognitive science, ``perceiving as predicting''~\cite{clark2014perceiving,clark2013whatever}, suggests that the human brain is not a simplified ``passive perceiving brain'', but rather an ``active predicting brain''~\cite{rao1999predictive,lee2003hierarchical} and that the cortical processing of sensory data plays the role of action-oriented inference that is driven by prior knowledge learned from endogenous experience. Likewise, in computer science, predictive learning, which models dynamic system states as predictions of future observations, has been of research interest for decades~\cite{pednault2000representation,eskandar1999dissociation,schaul2013better}. Specially, counterfactual predictions~\cite{levine2020offline} encode hypothesis-based inferences about what \textit{might} happen in the near or long-term future if the agent acts differently from prior experience~\cite{graves2020perception}. For example, we may efficiently improve car maneuvering skills using our sensory-motor experience by asking questions like ``what if steering the wheel slightly more towards the left''.

Motivated by the research mentioned above, We propose to use offline samples to learn a set of general value functions (GVFs~\cite{sutton2011horde}) that make counterfactual predictions from the visual inputs. GVFs provide a way to incorporate human prior knowledge of the task as cumulants~\cite{sutton2011horde} to learn meaningful predictions. We show that combining the offline learned counterfactual predictions (Fig. 1) with online force feedbacks allows efficient reinforcement learning given only a terminal reward signal. Our contributions are as below:

\begin{figure}
    	\setlength{\belowcaptionskip}{-10pt}
    	\begin{center}
    		\includegraphics[width=0.45\textwidth]{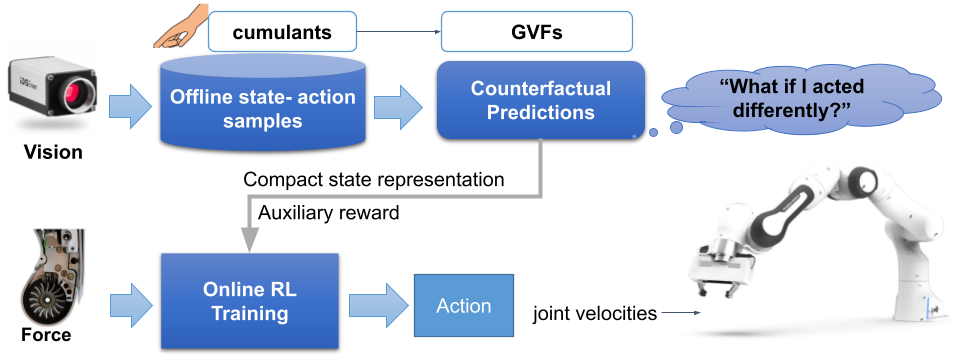} % Include the image placeholder.png
    		\caption{Overview of our method. We propose to encode visual inputs as counterfactual predictions using a set of general value functions (GVFs, see definitions in Section \ref{sec:gvf}). The learned predictions are then combined with force feedbacks in online policy learning, which form a compact representation that enables sample efficiency and provide auxiliary reward signals that guide online explorations towards contact-rich states.
    		}
    		\label{fig:design_overview}
    	\end{center}
\end{figure}	
 
 \begin{figure*}[tpb]
 	\setlength{\belowcaptionskip}{-10pt}
	\centering
	\includegraphics[width=0.95\textwidth]{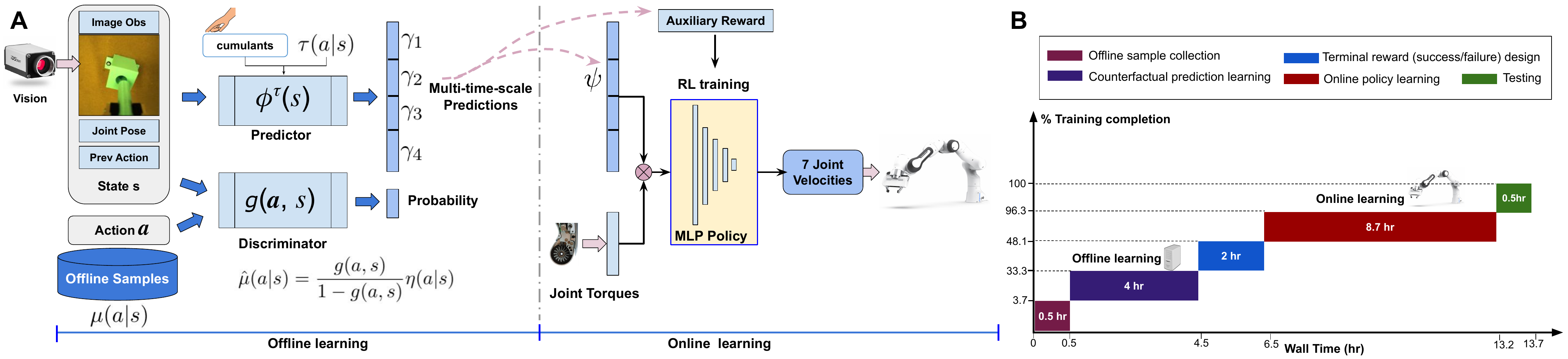}
	\caption{\textbf{A:} Work flow of our proposed offline and online learning approach. The offline training process aims to learn a counterfactual predictor $\phi^\tau(s) $ that maps high-dimensional ($3 \times 256 \times 256$) vision data to a compact vector based predictions considering multiple time scales, for example, $\gamma_{1}, \gamma_{2}, \gamma_{3} $ and $ \gamma_{4}$ \textbf{B:} Training timeline in real world testing of the exemplar task as described in Section \ref{sec_eval}.}
	\label{fig:training}
\end{figure*} 
    
 \begin{itemize}
\item We propose a reinforcement learning approach that combines offline and online learning, forming a practical solution to learn real-world robotic manipulation tasks. Specifically, we first learn to encode high-dimensional image states as compact counterfactual predictions using offline samples and then combine the learned predictions with low-dimensional force feedbacks during online training.
\item We show that the offline learned counterfactual predictions form a compact and informative representation, thus enabling sample efficiency and provide auxiliary reward signals to guide online explorations towards contact-rich states, thus enabling learning with only a terminal reward.
 \end{itemize}
 
Experiments both in simulation and real-world scenarios were conducted to support our claims.

\section{Related Works}

\label{sec:related}

This work is inspired by the following topics from both the reinforcement learning (RL) and robotics community.

\textbf{Real-world reinforcement learning of robotic tasks:} Addressing the various challenges of real-world robotic reinforcement learning has gained increasing interest recently~\cite{dulac2019challenges,zhu2020ingredients,mahmood2018setting, mahmood2018benchmarking}. There are a number of prior works tackling each of the challenges, for example, training efficiency~\cite{levine2013guided,gu2017deep}, generalization~\cite{zhang2018dissection,ponsen2009abstraction}, safer online explorations~\cite{mahmood2018setting,savinov2018episodic}, sim-to-real transfer~\cite{peng2018sim, chebotar2019closing} and learning from a sparse reward~\cite{vecerik2017leveraging,andrychowicz2017hindsight,fu2018variational}. However, very few works have been proposed to address the practicality of real-world robotic reinforcement learning. Recently, R3L~\cite{zhu2020ingredients} was proposed as a practical framework to learn tasks using a three-fingered robotic hand. It uses the VAE~\cite{kingma2013auto} encoder to form a compact state representation, utilizes VICE~\cite{fu2018variational} to provide rewards and enables training without resets. Compared to R3L: (1) We consider a more general problem setting of a 7-Dof robotic arm that learns a task using multimodal observations (vision and force). (2) We consider learning human prior knowledge as counterfactual predictions using offline data to accelerate online training since offline samples are easier to collect, and the offline learned human prior knowledge provides online exploration guidance given only a terminal reward.

% \textbf{Reinforcement learning with auxiliary tasks:} Our method shares similarities with auxiliary-task based approaches. Designing auxiliary tasks to facilitate the main task-learning has proven effective in many computer vision~\cite{zhang2014facial,mordan2018revisiting} and reinforcement learning tasks~\cite{jaderberg2016reinforcement,veeriah2019discovery,burda2018large}.  Among various approaches, general value functions (GVFs~\cite{sutton2011horde}) provide a flexible way to define the auxiliary tasks~\cite{jaderberg2016reinforcement,veeriah2019discovery} using human prior knowledge encoded in the form of cumulants.

\textbf{Learning a robotic insertion task:} Since we demonstrate our method’s advantages using a robotic insertion task, it is necessary to summarize existing works and discussion our distinctions under such task context. Learning a robotic insertion (a.k.a. peg-in-hole) task has been well studied~\cite{xu2019compare} in approaches including reinforcement learning ~\cite{lee2019making,vecerik2019practical,wu2019deep, ding2019transferable, luo2018deep, inoue2017deep} and contact-model based control\cite{hou2018learning,tang2016teach,song2014automated}. However, it is worth noting that most works require simplified assumptions that impede their practicality. 

\textit{Our work stands out by addressing the practical considerations as follows}. (1) Instead of simply assuming a near-perfect pre-insertion pose~\cite{ding2019transferable, luo2018deep, tang2016teach, song2014automated, inoue2017deep} at initialization, which is hard to define, the learned policy/controller should be able to start from an arbitrary grasping pose. Therefore, using visual feedback is important to guide robot motion starting from variable grasping poses. (2) Instead of relying on a hand-engineered controller~\cite{wu2019deep, ding2019transferable,lee2019making, vecerik2019practical} (such as admittance-Cartesian controller) which requires additional fine-tuning, the learned policy/controller should output actions that directly apply to robot joints. (3) Instead of relying on dense reward signals provided by a time-consuming perception system or calibrated instrumentation to obtain ground truth peg/slot poses~\cite{lee2019making}, the learning should be done using only a terminal reward (success/failure). (4) Instead of assuming a fixed peg position~\cite{lee2019making,vecerik2019practical,hou2018learning,tang2016teach} which does not consider a continuous process of peg feeding and inserting, as pointed out by~\cite{vecerik2019practical} in their limitation section, the learned policy should be robust to variable grasping poses.

\textbf{Predictive coding based representations:} Our method is similar to predictive coding based state representation learning methods~\cite{jin2019robot, hafner2019learning, hafner2019dream, okada2021dreaming, ma2020contrastive, yan2020learning} in that we encode high-dimensional visual observation as compact predictions. Our method can be viewed as more general way to learn predictions using general value functions that the cumulant in our method defines any meaningful predictions instead of only next state prediction in~\cite{okada2021dreaming, ma2020contrastive, yan2020learning}, and that the target policy in our method specifies predictions based any interesting behaviors instead of predictions only considering the environment dynamics.

\section{Method}
This section will introduce concepts of general value functions (GVFs), explain how to form counterfactual predictions using GVFs in the offline learning setting, and elaborate on how to use counterfactual predictions in a standard online reinforcement learning algorithm.

\subsection{Background: general value functions}
\label{sec:gvf}
Generally, a reinforcement learning agent aims to answer two questions from its environmental interactions~\cite{patterson2016}: (1) a prediction question in the form of a value function that predicts how much reward will be accumulated at state $s_{t}$ under a policy $\pi$; (2) a control-related question in the form of an optimal policy that concerns what action should be taken at state $s_{t}$ to achieve the maximum accumulated reward. We can view the reward in RL as a task-related question, for example, ``how well a robot circumvents pedestrians’’ in an obstacle avoidance task~\cite{jin2020mapless}. Then the value function provides the answer using predictions. As a result, a value function represents one aspect of the task or environment-related knowledge. Extending a value function to represent more general knowledge of the task or environment will derive the concept of general value functions (GVFs)~\cite{sutton2011horde}.

The core problem of GVFs is to learn answers to the prediction questions (GVF-questions). Like in a regular value function setting that a reward specifies a prediction question, GVF-questions are defined using different cumulates $c_{t}$ or pseudo rewards, which can be either hand designed (like, off-road measurement of an autonomous driving car~\cite{graves2021learning} or learned using knowledge discovery~\cite{veeriah2019discovery}). A general value function predicts the expected return of discounted cumulants $c_{t}$ by following a target policy $\tau$ according to a termination function $\gamma_{t}$ when at state $s_{t}$. Formally, a general value function is defined as:

\begin{equation}
\phi^\tau(s) = \E_{\tau}[\sum_{k=0}^{\infty} {(\prod_{j=0}^{k-1} {\gamma_{t+j+1}}) c_{t+k}} | s_t=s,a_t=a]
\label{eq_value}
\end{equation}
, where $\tau$ is the target policy for prediction, which is called a \textit{prediction demon}~\cite{sutton2011horde}, $\gamma_{t}$ is the discounting factor which relates to how a trajectory terminates with probability $1-\gamma_{t}$. Therefore, a GVF-question is defined by a tuple $(\tau, c_{t}, \gamma_{t})$. 

GVFs can be learned with off-the-shelf value function estimation methods based on temporal-difference (TD)~\cite{sutton2005temporal}. In practice, given a set of different cumulants $c_{t}$ and termination functions $\gamma_{t}$, GVFs learn many predictions defined by $c_{t}$ considering multi-temporal time scales defined by $gamma_{t}$ simultaneously.

\subsection{Learning Counterfactual predictions using offline samples}
\label{sec_count}

As described above, GVFs can be viewed as auxiliary tasks aiming to speed up learning the primary RL task. However, similar to learning a value function, learning GVFs in the real world remains challenging since it requires massive online robot-environment interactions. One practical approach is to use offline collected samples to learn to the predictions $\phi^\tau(s) $ in eq. (1), which we call counterfactual predictions. The adjective \textit{counterfactual}  means the predictions are based on a known prediction demon policy $\tau$, which is different from an unknown behaviour policy $\mu$ used to collect offline samples. For example, $\mu$ can be human demonstrations, robot random explorations or a mixture of both. Therefore, learning counterfactual predictions needs to address the \textit{distributional shift} issue.

\paragraph{Importance weighted temporal difference update} Suppose a general value function $\phi^\tau(s) $ is approximated using a deep neural network with parameters $\theta$. Learning a GVF $\phi^\tau(s) $ from samples collected with unknown behaviour policy $\mu (a|s)$ can be done using temporal different (TD) algorithms for off-policy learning. Specifically, an importance sampling correction term, as defined in~\cite{patterson2016} $\rho$, is used to address the difference between $\tau$ and $\mu$ as:
\begin{equation}
\rho = \frac{\tau (a|s)}{\mu (a|s)}
\end{equation}
, where $\rho$ is the probability density ratio of the target policy (prediction demon) $\tau$ over behavior policy $\mu$ given a state action pair $(s, a)$. $\tau$ is predefined as discussed later in Section \ref{sec_pre_defined} (a). Estimating $\mu (a|s)$ will be introduced later. 

Then given an offline dataset $\mathcal{D}$ collected with an known behavior policy $\mu$,  $\phi^\tau(s) $ can be learned by minimizing the importance weighted TD error $\delta$ as below:
 \begin{equation}
\mathcal{L} = \mathbb{E}_{(s,a) \sim \mathcal{D}}[\rho \delta ^{2}]
\end{equation}
, where $\delta=\phi^{\tau}(s;\theta)-y$ and $y$, which is similar to its use in a regular DQN algorithm~\cite{mnih2013playing}, is the target value computed by bootstrapping predictions of the currently estimated general value function $\phi^{\tau}(s;\hat{\theta})$ with the form below:
 \begin{equation}
y=\mathbb{E}_{s’ \sim P}[c_{t}+\gamma \phi^{\tau}(s’;\hat{\theta})]
\end{equation}

Then a GVF $\phi^\tau(s) $ is optimized by performing gradient descent as below:
 \begin{equation}
\nabla_{\theta} \mathcal{L} = \E_{(s,a) \sim \mathcal{D}}[{\rho} \delta \nabla_{\theta} \phi^{\tau}(s;\hat{\theta})]
\label{eq_gradient}
\end{equation}

\begingroup
\removelatexerror% Nullify \@latex@error
\begin{algorithm}[tpb]
	\SetAlgoLined
	\small
	\KwIn{ Offline samples $\mathcal{D}$, uniform distribution $\eta (a|s)$, cumulant function $c(s_{t}, a_{t}, s_{t+1})$, discounting factor $\gamma$, prediction demon policy $\tau (a|s)$}
	\KwResult{ Optimal weights ${\theta} ^{*}$ of a GVF $\phi ^{\tau} (s)$}
	%\textcolor[rgb]{0.14,0.36,0.73}{\textbf{Initialization}}\\
	Initialize $\phi^{\tau} (s)$, discriminator $g(a,s)$ and replay buffer $\mathcal{D}_{m}$\\
	%with random parameters ${\theta}_{0}$ and ${\psi}_{0}$ respectively.\\
	% Initialize replay memory $\mathcal{D}_{m}$\\
	\For{i=1:N}{
	$s_{0} \leftarrow $ sample the first state from $\mathcal{D}$\\
	    \For{t=0:T}{
	        \textcolor[rgb]{0.14,0.36,0.73}{\textbf{Optimize $\phi^{\tau}(s)$}}\\
	        $a_{t} \leftarrow $ obtain recorded $a_{t}$ from $\mathcal{D}$\\
	        $s_{t+1} \leftarrow$ obtain the next state from $\mathcal{D}$\\
	        Compute $c_{t}=c(s_{t}, a_{t}, s_{t+1})$\\
	        Estimate behavior density value $\hat{\mu}(a_{t}|s_{t})$ (eq. (8)).\\
	        Estimate importance sampling ratio $\rho _{t}=\frac{\tau(a_{t}|s_{t})}{\hat{\mu}(a_{t}|s_{t})}$\\
	        $\mathcal{D}_{m} \leftarrow$ Append transition $(s_{t}, a_{t}, \gamma, s_{t+1}, \rho_{t})$\\
	        Importance resampling according to Section \ref{sec_count}.\\
	        $A \leftarrow$ sample transitions $\{(s, a, c, \gamma, s', \rho)\}$ from $\mathcal{D}_{m}$ as described in Section \ref{sec_count}(a).\\
	        Compute target $y_{i}=c+\gamma \phi^{\tau}(s';\theta_{i-1})$\\
	        Compute one step TD error $\delta = \phi ^{\tau}(s;\theta_{i}) - y_{i}$\\
	        $\theta_{i+1} \leftarrow$ gradient descent using equation (6).\\
	        \textcolor[rgb]{0.14,0.36,0.73}{\textbf{Optimize $g(a, s)$}}\\
	        $B\leftarrow$ prepare balanced classification samples from $\mathcal{D}_{m}$ as described in  Section \ref{sec_count}(b).\\
	        $\psi_{i+1} \leftarrow$ update discriminator $g(a,s)$ with a binary cross entropy loss using samples in ${B}$.
	    }
	}
	\caption{  {Counterfactual prediction learning from offline samples collected with unknown $\mu (a|s)$}}
\end{algorithm}
\endgroup

In practice, the gradient update above with importance sampling ratios will have high variances. We apply the same importance resampling technique as defined in~\cite{schlegel2019importance} to address this issue. Specifically, suppose a replay buffer $D_{m}$ stores transitions $\{(s, a, c, \gamma, s', \rho)\}$ with size $N$. An importance resampling process makes the sampling probability $p_{i}$ of each sample $i=1, ..., N$ proportional to its $\rho_{i}$ as $p_{i}\frac{\rho_{i}}{\sum^{N}_{j=1} \rho _{j}}$. Therefore, a new replay buffer $\mathcal{D}_{\rho}$ is constructed by the resampling process. Moreover, suppose a mini-batch A is sampled from $\mathcal{D}_{\rho}$, to further reduce the update variances caused by $\rho$, we replace it by computing the average $\bar{\rho}$ in the mini-batch. In summary, the update gradient in eq. \ref{eq_gradient} is computed using:
 \begin{equation}
\nabla_{\theta} \mathcal{L} = \E_{(s,a) \sim \mathcal{D}_{\rho}}[\bar{\rho} \delta \nabla_{\theta} \phi^{\tau}(s;\hat{\theta})]
\label{eq_gradient2}
\end{equation}

\paragraph{Estimating the unknown behaviour policy $\mu$}

The unknown behaviour policy $\mu (a|s)$ in eq. (2) is estimated using a density ratio trick where a discriminator $g(a, s)$ is used to distinguish samples from $\mu (a|s)$ and an intermediate probability density function $\eta (a|s)$. We choose $\eta (a|s)$ as a uniform distribution so that the discriminator $g(a,s)$ can well distinguish samples from $\mu (a|s)$ over all possible actions at state $s$. Specifically, we sample a minibatch $B$ from offline dataset $\mathcal{D}$ and mark the samples with label $z=1$, which means a state-action pair $(s, a)$ is generated by $\mu (a|s)$. Then we randomly set a half size of the samples in $B$ with label $z=0$, which means the sample belongs to $\eta (a|s)$ by generating uniform actions. Then the discriminator is trained by a cross-entropy loss to differentiate samples with class labels $z=1$ and $z=0$.  By applying the density ration trick and assuming $p(z=1)=p(z=0)$, we have:
\begin{equation}
    \begin{split}
        \frac{\mu(a|s)}{\eta(a|s)} & = \frac{p(a|s,z=1)}{p(a|s,z=0)} = \frac{p(z=1|a,s)/p(z=1)}{p(z=0|a,s)/p(z=0)} \\
            & = \frac{p(z=1|a,s)}{p(z=0|a,s)} = \frac{g(a,s)}{1 - g(a,s)}
    \end{split}
\label{eq_condition_density_ratio}
\end{equation}
Therefore, the probability density value of behavior policy $\mu$ given a state action pair $(s,a)$ is estimated as:
\begin{equation}
\hat{\mu} (a|s)=\frac{g(a,s)}{1 - g(a,s)}
{\eta (a|s)}
\end{equation}

Algorithm 1 concludes learning counterfactual predictions $\phi^{\tau}$ using offline samples.

\paragraph{Learning multi-temporal time scale predictions}
As shown in~\cite{schaul2013better,graves2020perception}, multi-temporal-scale predictions help learning more efficiently. $\gamma$ controls a temporal time horizon of predictions.  In experiments, we use predictions on four different time scales for the near and long-term future with $\gamma=[0,0.5,0.9,0.95]$. Specifically, we run Algorithm 1 for each $\gamma$ value and stack the predictions of $\phi^{\tau}$ as the counterfactual predictions spanning multi-temporal time horizons. 

\subsection {Counterfactual predictions for real-world task learning}
\label{sec_pre_defined}
Let us go back to our problem setting---learning robotic manipulation tasks with multi-modal sensory inputs (vision and force) with a terminal reward $z_{t}$ for practical considerations as discussed in Section \ref{sec:intro}. 

Suppose the observations at time t are image $I_{t}$, robot joint angles $J_{t}$, and joint force feedback $F_{t}$. We aim to learn a policy that maps the above observations to robot actions $a_{t}$, which are the joint velocities. Our strategy is to learn counterfactual predictions to encode visual observations using offline samples. Then we combine the predictions with force feedback in the online training. 

We apply this strategy to address the multi-modality issue not only for dimension reduction but also for the consideration that collecting force-sensitive samples offline using either a scripted policy or human demonstrations is tedious in real-world settings. On the other hand, as shown in~\cite{wu2019deep, ding2019transferable, luo2018deep}, if the robot learns to touch the object using vision, contact-rich skills will be much easier to learn using standard off-policy learning methods in an end-to-end manner.

\paragraph{Defining cumulants and prediction demon policy}
\label{sec:cumulant_define}
We define the cumulants $c_{t}$ as a 6-dimensional vector representing the relative pose between the end-effector and the target. In experiments, the cumulants are defined by human kinesthetic teaching the robot where the target is without actually inserting the peg. The main RL task is to learn a policy that moves the peg towards the slot and inserts it. 

For the prediction demon policy $\tau$, we use a previous-action-conditioned policy  $\tau(a_t|s_t, a_{t-1})=\mathcal{N}(a_{t-1}, \Sigma)$, where $\Sigma=0.0025I$ is a diagonal covariance matrix. It means ``continuing with the previous action while allowing some randomness''. Thus the predictions encode how task spatial information will evolve if the agent continues on previous actions. These predictions form corrective signals for decision-making. 

\paragraph{Offline learning}
To learn $\phi^{\tau}(s)$ using offline samples, as shown in Fig. 2A, we define the state as $s_{t}=[I_{t}, J_{t}, a_{t-1}]$, where $a_{t-1}$ is the previous action and initialized using a zero vector at $t=0$. The GVF $\phi^{\tau}(s)$ is parameterized using a deep neural network with convolutional layers to encode $I_{t}$ and a concatenate operation to incorporate $J_{t}$ and $a_{t-1}$.  $\phi^{\tau}(s)$ then outputs a 24-dimensional vector $\psi_{t} \in \mathbb{R}^{24}$ representing the counterfactual predictions of each of the six pose elements considering multi-temporal time scales with $\gamma=[0,0.5,0.9,0.95]$. Let us write:
\begin{equation}
    \psi _{t} = \phi^{\tau}(I_{t}, J_{t}, a_{t-1})
\end{equation}

After collecting offline samples to construct dataset $\mathcal{D}$, Algorithm 1 is used to optimize $\phi^{\tau}$.

\paragraph{Online Training}
During online training, as shown in Fig. 2A, we construct a new state $x_{t}=[\psi _{t}, F_{t}]$ by incorporating force feedbacks. Suppose the robot is a 7 Dof manipulator that outputs 7-dimensional force feedback. Then the new state $x_{t} \in \mathbb{R}^{31}$ is a 31-dimensional vector. Thus, as shown in Fig. 2A, we can use linear layers to parameterize the policy which outputs continous 7-dimensional joint velocities.

Although the state $x_{t}$ is compact, given only a terminal reward $z_{t}$, it is still challenging to train the policy. We propose to use the counterfactual predictions $\psi _{t}$ as an auxiliary reward to guide online policy learning.  Specifically, the auxiliary reward $\alpha _{t}$ is defined as $\alpha_{t}=exp(-\lambda \norm{\psi _{t}})$, where $\lambda$ is a scaling coefficient. $\alpha_{t}$ then encodes how the relative pose between the robot end-effector and the target will evolve if the robot continues with the previous action $a_{t-1}$. So $\alpha_{t}$  will guide the online exploration to touch the target towards contact-rich interactions. Then the augmented reward function for training is:

\begin{equation}
r_{t}=
\begin{cases}
z_{t} & \text{ if succeed,}\\
\alpha _{t} & \text{ otherwise}
\end{cases}
\end{equation}

\begin{figure*}
	\setlength{\belowcaptionskip}{-10pt}
	\centering
	\includegraphics[width=0.95\textwidth]{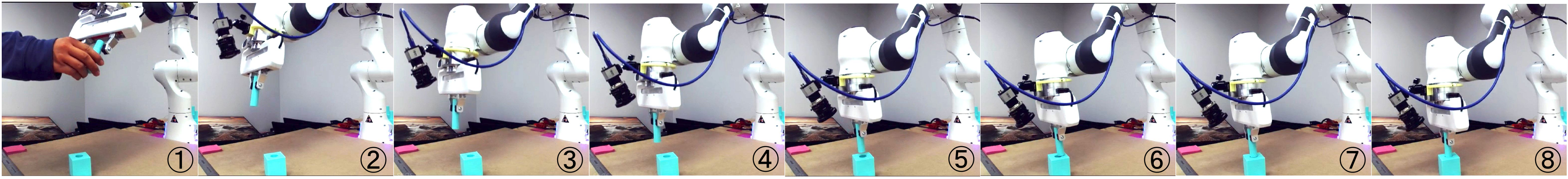}
	\caption{visualization of a testing trial, which includes human randomly feeding the peg, robot grasping the peg, and inserting the peg.}
	\label{fig:training}
\end{figure*}

\begin{figure*}[tbp]
\setlength{\belowcaptionskip}{-10pt}
	\centering
	\begin{tabular}{@{}p{44mm}@{}p{44mm}@{}p{44mm}@{}p{44mm}}
	    \includegraphics[height=3.5cm]{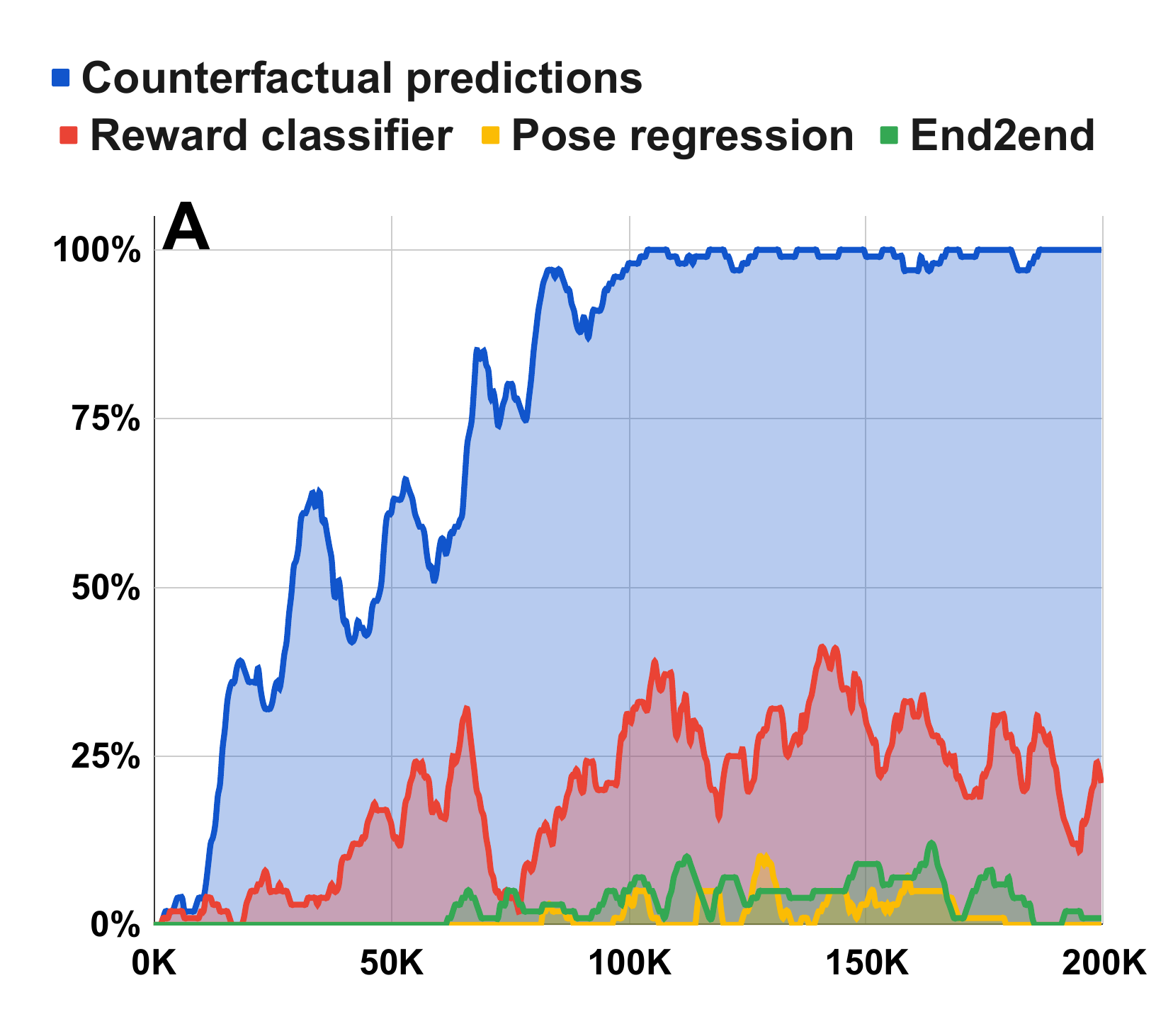}
		&{\includegraphics[height=3.5cm]{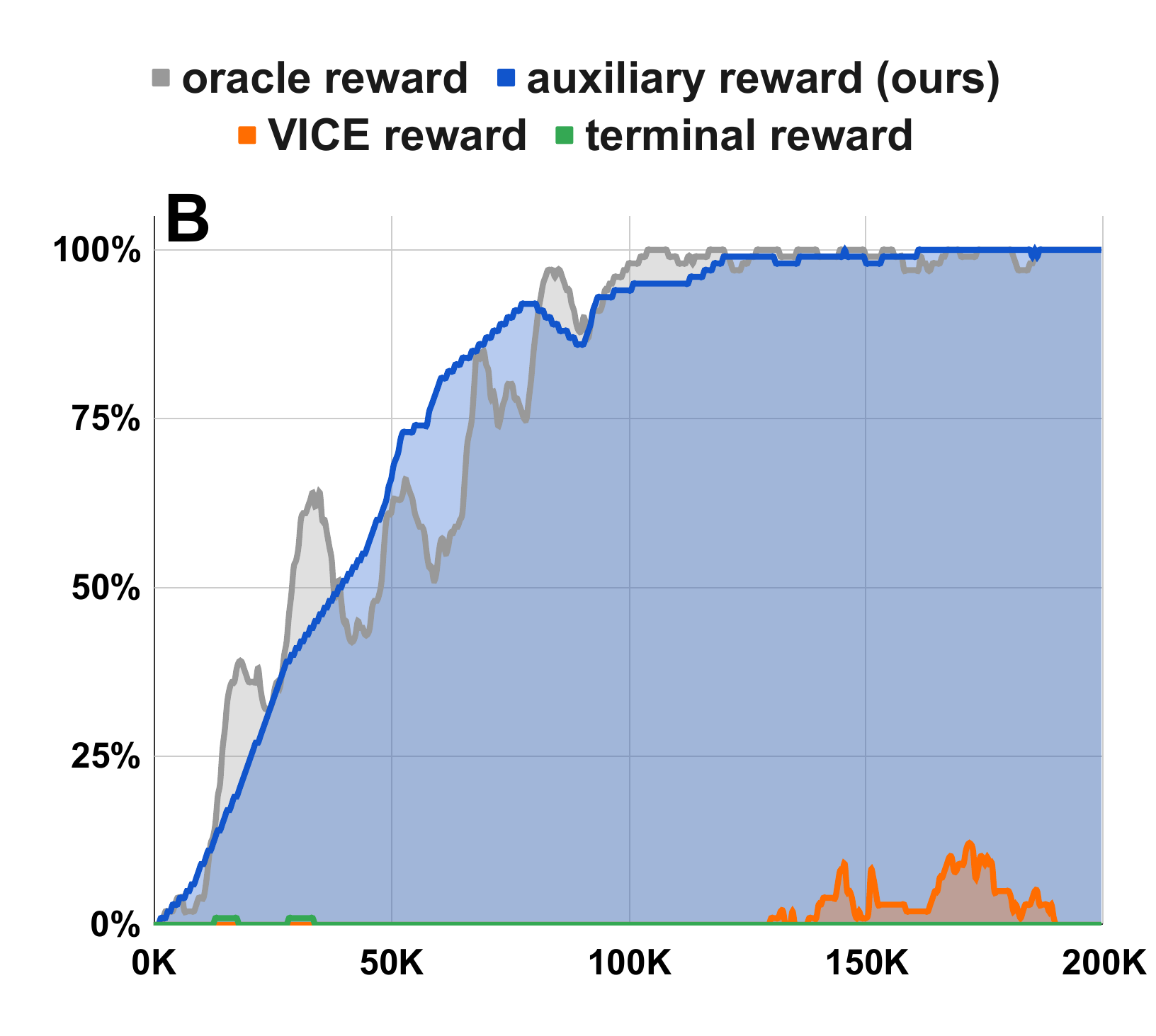}}
		&{\includegraphics[height=3.5cm]{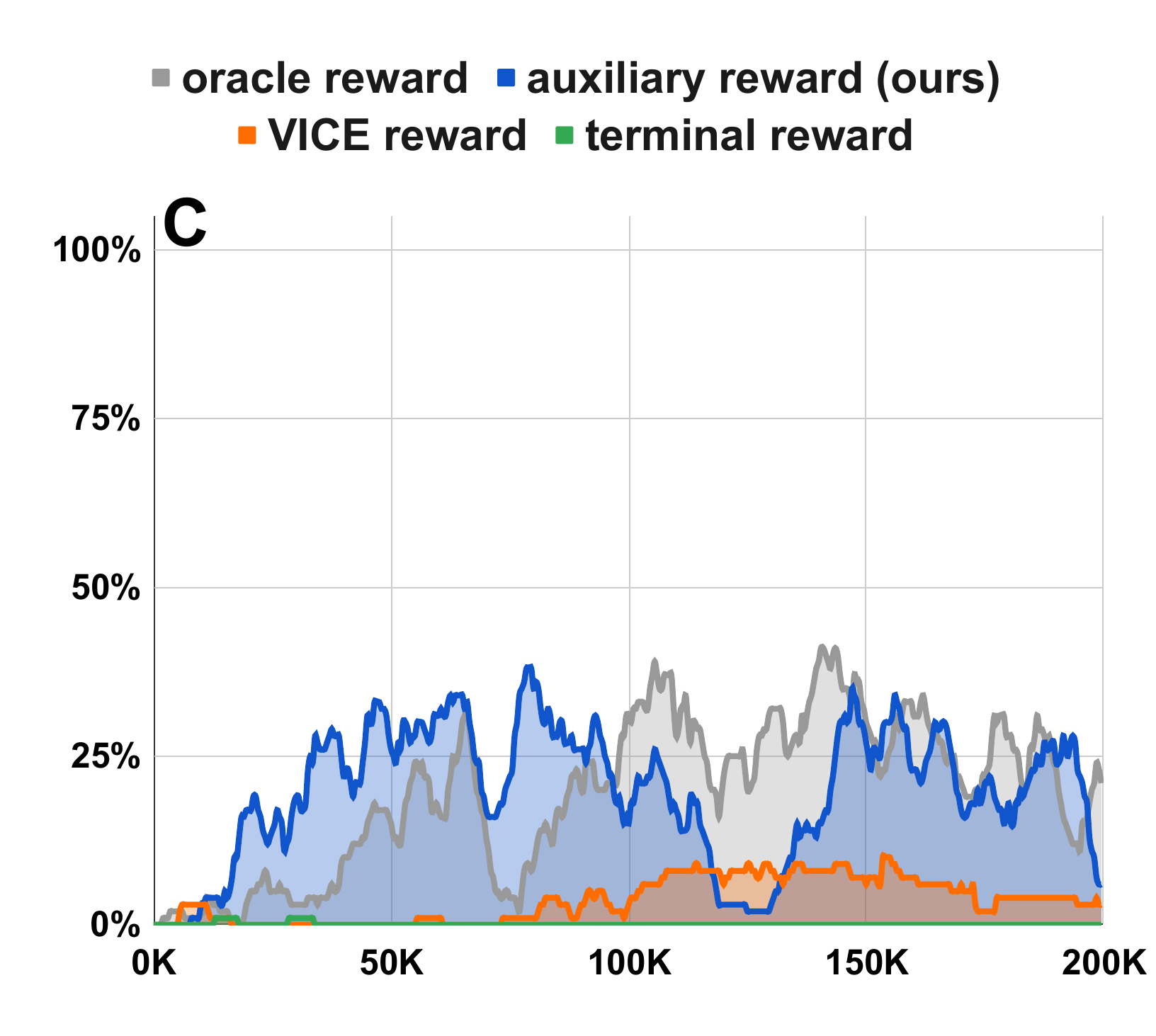}}
		&{\includegraphics[height=3.5cm]{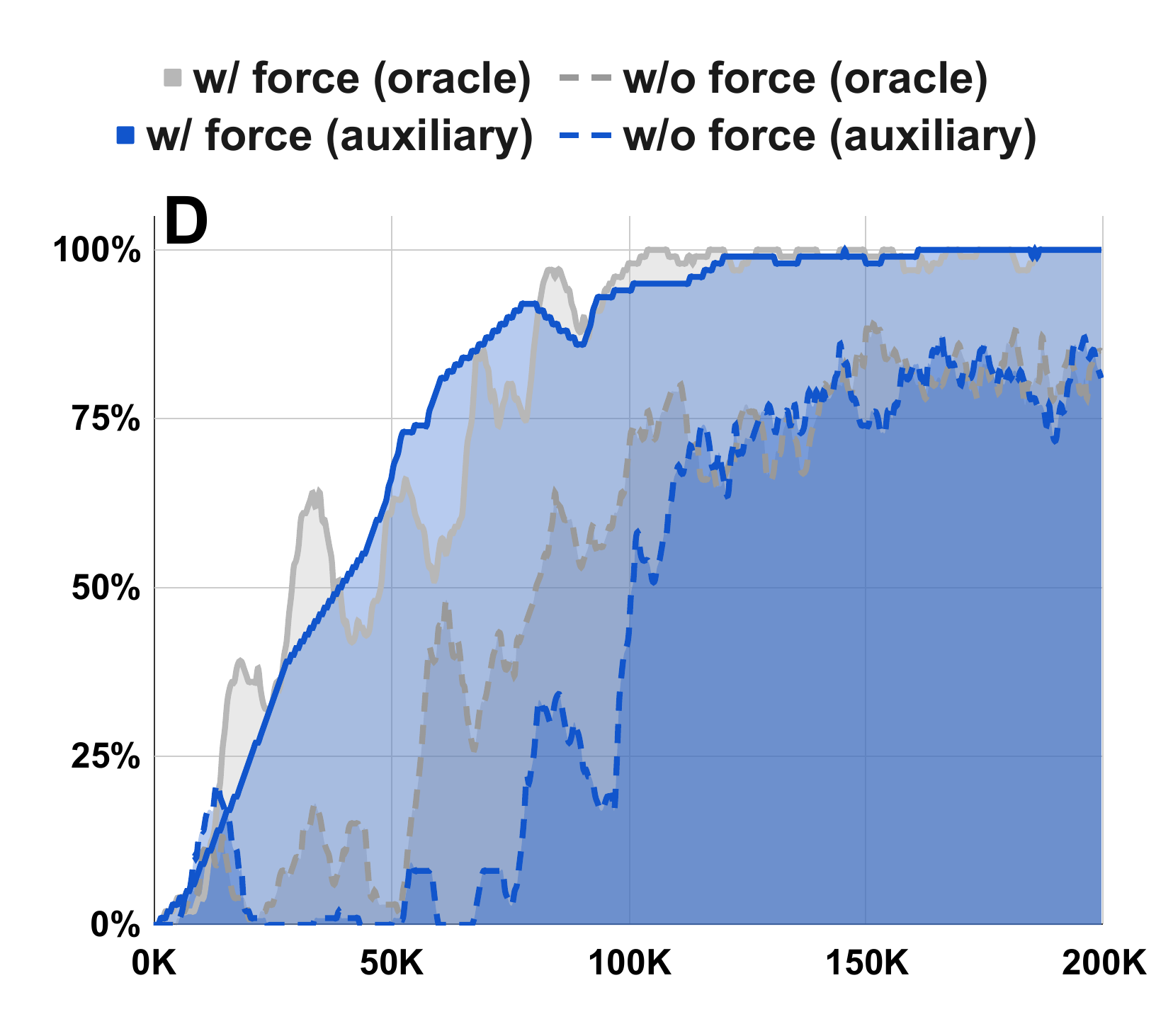}}
	\end{tabular}
\caption{Evaluation results in simulator. The horizontal axis shows the number of samples used during online training. The vertical axis shows success rate of recent 100 trials. \textbf{A}: State representation performance of our proposed counterfactual predictions (blue) compared to three baselines given a fine-tuned oracle reward. \textbf{B and C:} Effectiveness of our counterfactual prediction based auxiliary reward (blue) compared to three baseline rewards under the same state representations (B: counterfactual predictions as the state representation. C: using the second last layer of a reward classifier~\cite{vecerik2019practical,vecerik2017leveraging} as the state representation.) \textbf{D:} Effectiveness of combining counterfactual predictions with (solid line) and without (dashed line) force feedback under a fine-tuned oracle reward and an  auxiliary reward. }
	\label{fig:as}
\end{figure*}

Given the new state representation $x_{t}$ and the augmented reward function $r_{t}$, any off-the-shelf policy optimization algorithms~\cite{haarnoja2018soft,lillicrap2015continuous} can be used during online training.

\section{Evaluation}
\label{sec_eval}
% Results show our method significantly outperforms other state representation methods. Both results show our proposed auxiliary reward is competent to the fine-tuned oracle reward and significantly outperforms other rewards. Results under a fine-tuned oracle reward and our proposed auxiliary reward all show combining force feedback in online training will improve the performance. 

We aim to evaluate on four objectives: (1) the representation performance of counterfactual predictions; (2) the effectiveness of an auxiliary reward using counterfactual predictions; (3) the effectiveness of combining counterfactual predictions with online force feedback; (4) the viability of our method in a real-world task setting regarding how long the training takes and how hard to make the training work (such as if the training requires tediously designed perception pipelines or calibrated instrumentation).

We use a peg-in-hole task for evaluation. Experiments were conducted in a real-world and simulation scenario. As shown in Fig. 5, the experimental setup includes an eye-in-hand camera and a 7 Dof Franka Emika Panda robot. Each initialization will randomize the poses of robot joints, grasping peg and the slot. In real-world experiments, the slot pose was fixed during training due to technical limitations to reset the slot pose automatically. Nevertheless, because the robot acquired force-sensitive probing skills (Fig. 8), we found the learned policy is relatively robust to a slot displacement $<$ 1cm.

\begin{figure}[h]
	\setlength{\belowcaptionskip}{-10pt}
	\begin{center}
		\includegraphics[width=0.3\textwidth]{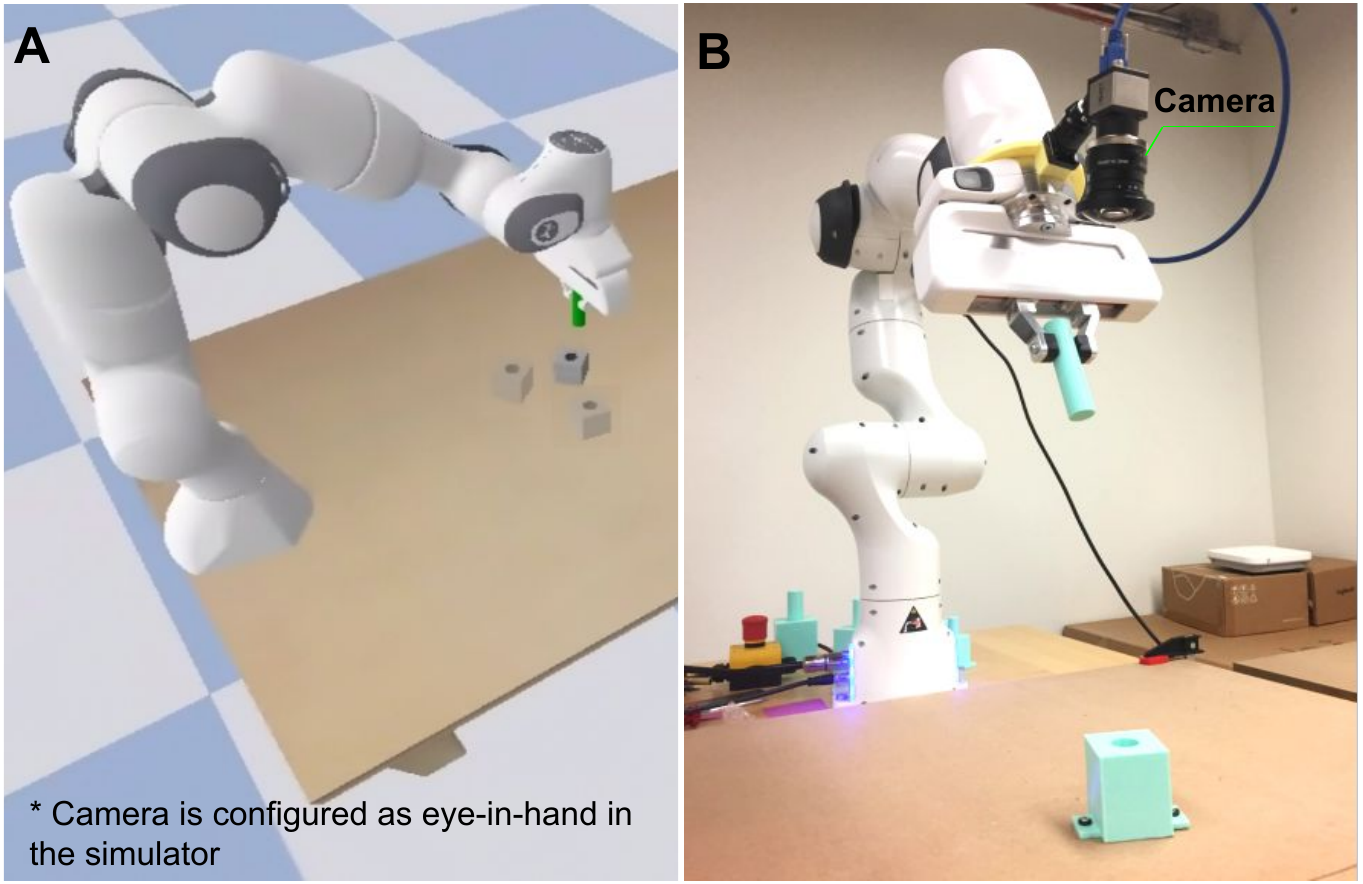} 
		\caption{Experimental setup. \textbf{A:} Simulation. \textbf{B:} Real world.}
		\label{fig:a}
	\end{center}
\end{figure}

\subsection{Evaluation in simulator}
The simulator provides an ideal environment for quantitative comparison between our method and different baselines since the ground truth peg/slot poses are easy to obtain.

\paragraph{State representation performance of counterfactual predictions}
We evaluate the state representation performance of counterfactual predictions compared to different baselines. For all the representation methods evaluated, we give a fine-tuned oracle reward, as defined in~\cite{lee2019making}, that provides a staged dense reward to guide the peg insertion using ground truth peg/slot poses. We compare our method with three representation baselines. (1) A reward classifier baseline used in~\cite{vecerik2019practical,vecerik2017leveraging}, which trains a classifier to distinguish success/failure images and use the second last layer as the image representation. (2) A pose regression baseline used in ~\cite{zeng2018robotic}, which trains a regressor that predicts the peg pose given an image input and uses the second last layer as the image representation. (3) An end-to-end baseline that uses convolutional layers to encode the image without any pre-training. For a fair comparison, all baselines output a vector with the same dimension as our counterfactual predictions. Therefore, the linear policy for all the state representation methods shares the same architecture. Results are shown in Fig. 4A, which indicates counterfactual predictions significantly outperform other state representation methods.

\paragraph{Counterfactual predictions as an auxiliary reward}
We evaluate the effectiveness of the auxiliary reward based on counterfactual predictions. We study the effectiveness by comparing our proposed auxiliary reward with (1) the fine-tuned oracle reward, (2) a VICE reward~\cite{zhu2020ingredients, fu2018variational} which computes the probability of success using a success/failure classifier’s output and (3) a terminal reward (success/failure). Here, we set the terminal success reward as $+10$ and the failure penalty as $-10$.

We compare different rewards under two scenarios using different state representations: (1) our counterfactual predictions and (2) a reward classifier-based~\cite{vecerik2019practical,vecerik2017leveraging} state representation as described above. Results are shown in Fig. 4B and C, which indicate our proposed auxiliary reward is competent with an oracle reward and significantly outperforms a VICE reward and the terminal reward.

\paragraph{Ablation study on multimodal sensory inputs}
Next, we study the effectiveness of combining counterfactual predictions with online force feedbacks by online training with and without force feedback. Results are shown in Fig. 4D, which indicates our approach that combines force feedback during online training improves the learning performance.

%The results in the simulator demonstrate the effectiveness of using counterfactual predictions as a state representation, as an auxiliary reward and as a combination with force feedback. These results give us clues in the real-world evaluation.

\subsection{Evaluation on a real-world robot}
The total training time of our real-world experiment is about 13.2 hrs which contains 4.5 hrs offline and 8.7 hrs online training.

\begin{figure}[h]
	\setlength{\belowcaptionskip}{-10pt}
	\begin{center}
		\includegraphics[width=0.3\textwidth]{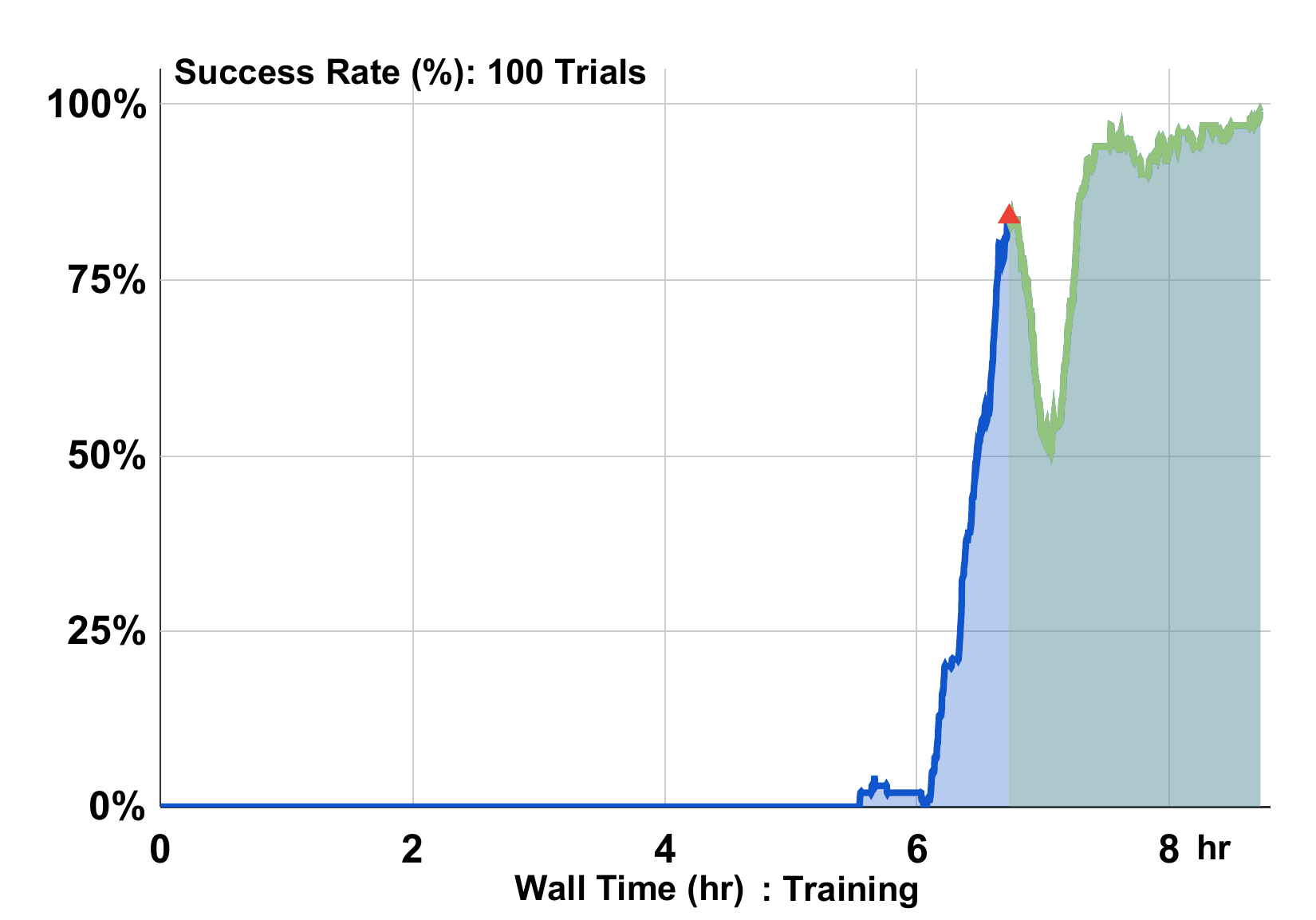} 
		\caption{Training curve of the online learning. The horizontal axis is the time duration and the vertical axis is the success rate of the most recent 100 trials. The red triangle and the shaded green region indicates human randomly changing the grasping pose in order to make the learned policy robust to variable peg grasping pose.}
		\label{fig:a}
	\end{center}
\end{figure}

\paragraph{Offline training}
We collected 40K offline samples by 400 runs which took 0.5 hr including the human teaching time. Samples were collected using a position-based visual servoing (PBVS~\cite{chaumette2007visual}) controller and a random action controller, with each contributed 20k samples.

\paragraph{Online training}
Soft-actor-critic (SAC~\cite{haarnoja2018soft}) was used in online training with a 30Hz control frequency that matches the frame rate of our vision sensor. Joint velocity actions were directly sent to the \textit{franka\_ROS} controller at 30Hz, and the internal controller of the Franka arm runs at 1000Hz, controlling robot motions with default and unknown parameters without fine-tuning. We observe that our learned policy can generate smooth motions.

% \begin{figure}[h]
% 	\setlength{\belowcaptionskip}{-10pt}
% 	\begin{center}
% 		\includegraphics[width=0.3\textwidth]{figure/fig7_v2.pdf} 
% 		\caption{The sfg}
		
% 		\label{fig:a21}
% 	\end{center}
% \end{figure}

The terminal reward is provided by detecting the success/failure status given an image. Similar to~\cite{vecerik2019practical}, we used two success/failure classifiers independently trained to avoid false-positive results of the terminal reward. 

The training curve of the online learning using SAC~\cite{haarnoja2018soft} is shown in Fig. 6. Note that, unlike most methods assuming a fixed peg position~\cite{lee2019making,vecerik2019practical,hou2018learning,tang2016teach}, in order to make the learned policy robust to variable peg grasping pose, we started to randomly change the grasping pose when the success rate reached $84\%$ after 6.8 hrs of training.

\begin{figure}[h]
	\setlength{\belowcaptionskip}{-10pt}
	\begin{center}
		\includegraphics[width=0.3\textwidth]{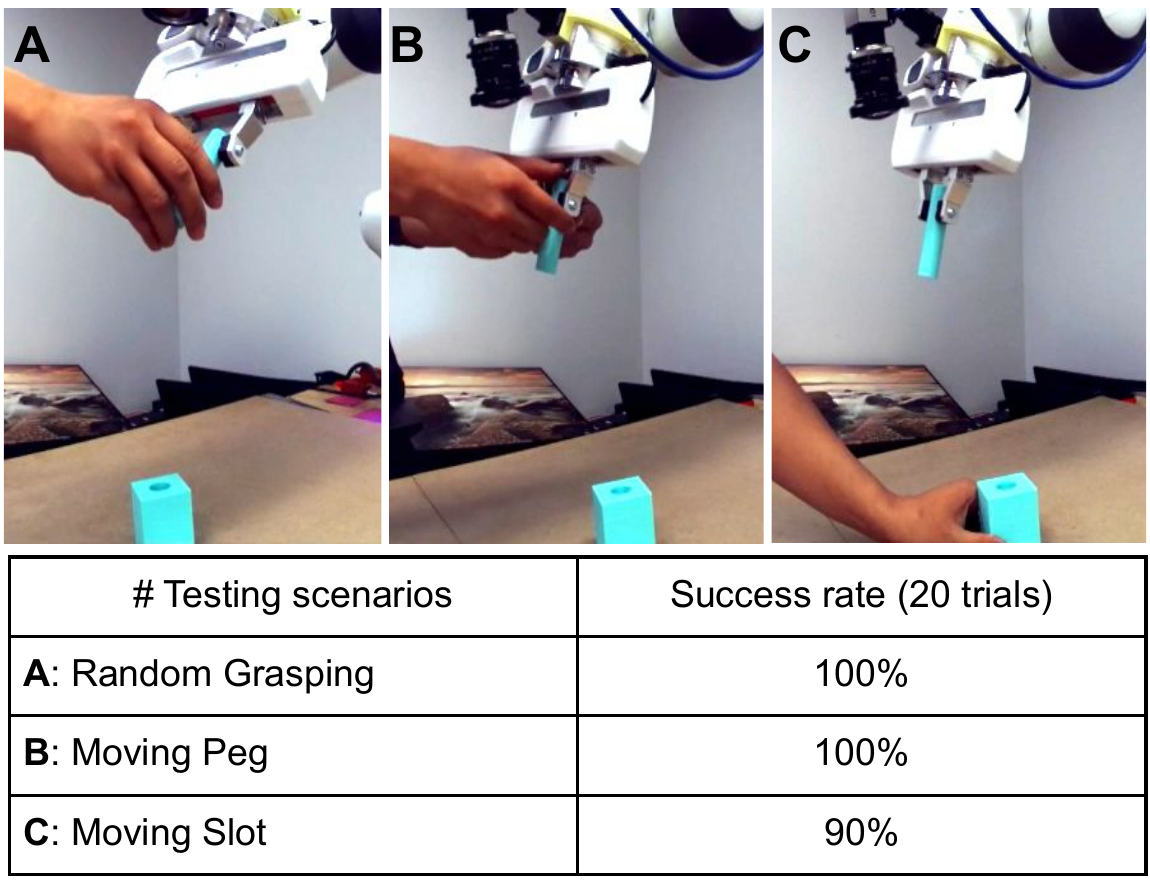} 
		\caption{Testing results under three changing scenarios.}
		
		\label{fig:a}
	\end{center}
\end{figure}
\begin{figure}
	\setlength{\belowcaptionskip}{-10pt}
	\begin{center}
		\includegraphics[width=0.35\textwidth]{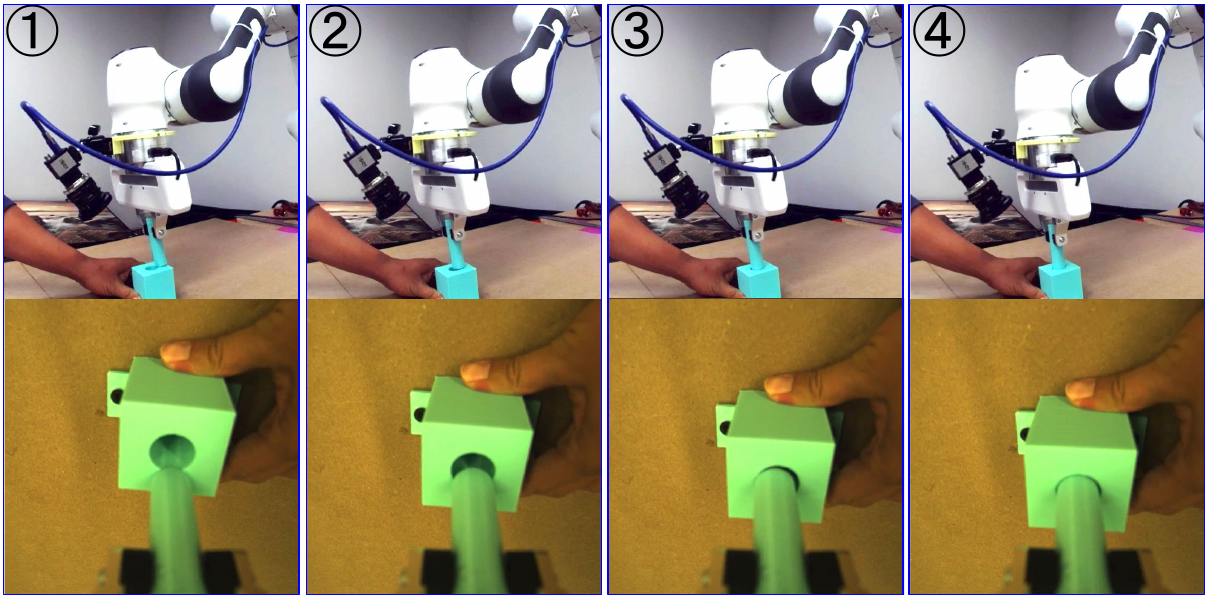}
		\caption{The force-sensitive probing behavior learned by the robot. Without an additional wrist F/T sensor, the learned policy can adjust its joint velocity outputs according to joint force feedback when human randomly moves the slot . \textbf{Up:} Scene viewed from a separate camera. \textbf{Down:} Image view from the wrist mounted camera used in robot control.}
		
		\label{fig:a}
	\end{center}
\end{figure}

\paragraph{Testing}
The learned policy was then evaluated under scenarios of random  grasping (Fig. 7A), moving the peg (Fig. 7B) and slot (Fig. 7C) in a continuous testing process (Fig. 5) with 20 trials. Results (Fig. 8) show our proposed method’s practicality for training a real-world robotic RL agent to acquire fine manipulation skills in about half a day, without hand-engineered perception systems or calibrated instrumentation.

\section{Conclusion}
\label{sec:conclusion}
We propose an offline counterfactual prediction learning method to address challenges in real-world RL considering a general robotic manipulation task setting that maps both vision and force observations directly to robot joint velocities. The key insight that makes our method work is the use of general value functions to encode high-dimensional vision data as compact predictions spanning multiple time scales. We call it counterfactual since the predictions are learned using offline collected samples by a target policy which is different from the main task policy during online training. In summary, our method uses an offline-online training scenario combined with off-policy learning.   

Although we demonstrate the effectiveness of GVF-questions which are hand crafted to encode spatial information (relative pose in Section \ref{sec:cumulant_define}), one may wonder how to generalize our method without a hand crafted cumulant since there should be more meaningful representations that can be learned from the offline collected data. This remains an open question and further research on automatic knowledge discovery ~\cite{veeriah2019discovery} from data is worth being explored.

	\addtolength{\textheight}{-0 cm}   % This command serves to balance the column lengths
	
	%%%%%%%%%%%%%%%%%%%%%%%%%%%%%%%%%%%%%%%%%%%%%%%%%%%%%%%%%%%%%%%%%%%%%%%%%%%%%%%%

	%%%%%%%%%%%%%%%%%%%%%%%%%%%%%%%%%%%%%%%%%%%%%%%%%%%%%%%%%%%%%%%%%%%%%%%%%%%%%%%%

	%%%%%%%%%%%%%%%%%%%%%%%%%%%%%%%%%%%%%%%%%%%%%%%%%%%%%%%%%%%%%%%%%%%%%%%%%%%%%%%%

	%%%%%%%%%%%%%%%%%%%%%%%%%%%%%%%%%%%%%%%%%%%%%%%%%%%%%%%%%%%%%%%%%%%%%%%%%%%%%%%%
	
% 	\clearpage
	
	\bibliographystyle{IEEEtran}
	\bibliography{IEEEabrv,IEEEexample}
	
\end{document}